\documentclass[letterpaper]{article} % DO NOT CHANGE THIS
\usepackage{aaai23}  % DO NOT CHANGE THIS
\usepackage{times}  % DO NOT CHANGE THIS
\usepackage{helvet}  % DO NOT CHANGE THIS
\usepackage{courier}  % DO NOT CHANGE THIS
\usepackage[hyphens]{url}  % DO NOT CHANGE THIS
\usepackage{graphicx} % DO NOT CHANGE THIS
\usepackage{natbib}  % DO NOT CHANGE THIS AND DO NOT ADD ANY OPTIONS TO IT
\usepackage{caption} % DO NOT CHANGE THIS AND DO NOT ADD ANY OPTIONS TO IT
\usepackage{algorithm}
\usepackage{algorithmic}
\usepackage{amsfonts}
\usepackage{subcaption}
\usepackage{multicol}

\newcommand{\Tau}{\mathcal{T}}
\usepackage{amsmath}
\usepackage{amssymb}
\usepackage{booktabs}
\usepackage{xcolor}
\usepackage{xspace}

\newcommand*{\eg}{{\em e.g.~}\@\xspace}
\newcommand*{\ie}{{\em i.e.~}\@\xspace}
\usepackage{newfloat}
\usepackage{listings}
\DeclareCaptionStyle{ruled}{labelfont=normalfont,labelsep=colon,strut=off} % DO NOT CHANGE THIS
\floatstyle{ruled}
\newfloat{listing}{tb}{lst}{}
\floatname{listing}{Listing}
\title{On the Challenges of using Reinforcement Learning in Precision Drug Dosing:\\Delay and Prolongedness of Action Effects}
\author{
    Sumana Basu\textsuperscript{\rm 1,\rm 2},
    Marc-André Legault\textsuperscript{\rm 1,\rm 2},
    Adriana Romero-Soriano\textsuperscript{\rm 1,\rm 2,\rm 3},
    Doina Precup\textsuperscript{\rm 1,\rm 2}
}
\affiliations{
    \textsuperscript{\rm 1} McGill University,
    \textsuperscript{\rm 2} Mila,
    \textsuperscript{\rm 3} Meta AI \\
    sumana.basu@mail.mcgill.ca,
    marc-andre.legault@mcgill.ca,
    adriana.romsor@gmail.com,
    dprecup@cs.mcgill.ca
}

\usepackage{bibentry}
\begin{document}

\maketitle

\begin{abstract}
Drug dosing is an important application of AI, which can be formulated as a Reinforcement Learning (RL) problem. In this paper, we identify two major challenges of using RL for drug dosing: \textit{delayed} and \textit{prolonged} effects of administering medications, which break the Markov assumption of the RL framework. We focus on prolongedness and define \textit{PAE-POMDP (Prolonged Action Effect-Partially Observable Markov Decision Process)}, a subclass of POMDPs in which the Markov assumption does not hold specifically due to prolonged effects of actions. Motivated by the pharmacology literature, we propose a simple and effective approach to converting drug dosing PAE-POMDPs into MDPs, enabling the use of the existing RL algorithms to solve such problems.

We validate the proposed approach on a toy task, and a challenging glucose control task, for which we devise a clinically-inspired reward function. Our results demonstrate that: (1) the proposed method to restore the Markov assumption leads to significant improvements over a vanilla baseline; 
(2) the approach is competitive with recurrent policies which may inherently capture the prolonged effect of actions; (3) it is remarkably more time and memory efficient than the recurrent baseline and hence more suitable for real-time dosing control systems; and (4) it exhibits favourable qualitative behavior in our policy analysis.\looseness-1
\end{abstract}

\section{Introduction}
Drug dosing plays an important role in human health-- \eg 
%Drug dosing is an important challenge under various circumstances-- \eg 
individuals with type 1 diabetes require regular insulin injections to manage their blood glucose levels, intensive care patients require continuous monitoring and administration of drugs, optimal doses of anaesthesia are required during operative procedures, etc. Optimal drug dosing is most important in cases where the therapeutic window is narrow, meaning that small deviations from the therapeutic range of drug concentration may lead to serious clinical complications~\cite{fda2015, maxfield2021}. These problems are compounded by idiosyncratic differences in the dynamics of drug absorption, distribution, metabolism or excretion (collectively referred to as pharmacokinetics) and drug sensitivity. Therefore, one important goal of precision medicine is to tailor patient care while accounting for individual characteristics, and  developing algorithmic solutions for drug dosing is a contribution towards that broader goal.\looseness-1

\begin{figure}
  \centering
 \includegraphics[width=0.75\columnwidth]{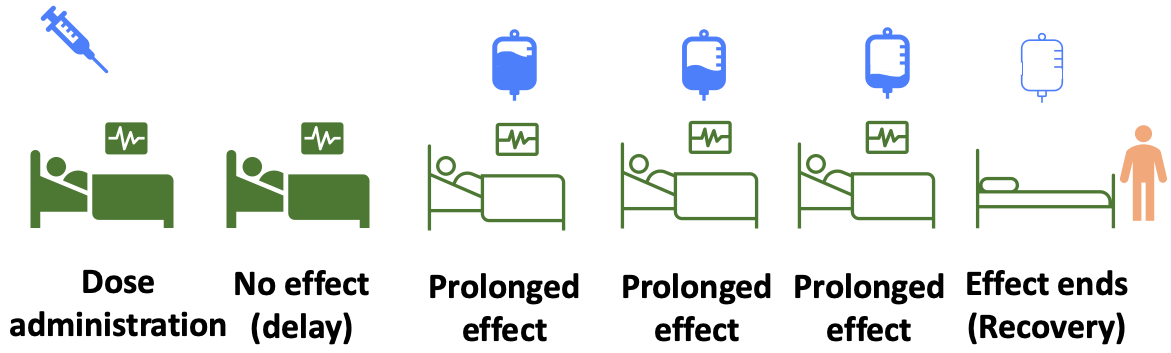}
  \caption{\textbf{Delayed and prolonged effects of drugs.} Drug is only administered at the first step, but no immediate effect is observed. After the initial delay, the initial dose keeps altering the patient status for a while, even in the absence of additional dosage. The duration of delay and prolongedness is individual specific.\vspace{-0.2cm}\looseness-1}
  \label{fig:motivation}
\end{figure}

Reinforcement learning (RL) offers a framework to account for individual characteristics and automatically derive personalized treatment policies in line with the objective of precision medicine ~\cite{ribba2020}. However, RL-based algorithms cannot be applied off-the-shelf to tackle precision dosing since all drugs are known to have a \emph{delayed} and \emph{prolonged} effect from the point of medication \cite{nick2018} (see Figure~\ref{fig:motivation}). The delay is attributed to the time it takes for the drug to distribute to the target site, bind to the receptor and finally to change physiological substances \emph{before} its response can be observed. This can vary between minutes to hours or even longer \cite{nick2018}. The prolongedness is due to individual variation in pharmacokinetics \cite{vogen2010}. In this delayed and prolonged action effects scenario, the future depends on the previous drug dosages and their effect, and therefore the Markov assumption usually made by RL algorithms no longer holds. Although RL has been applied to address drug dosing problems such as controlling glucose levels for closed loop artificial pancreas~\cite{tejedor2020}, the violation of the Markov assumption is in this case not only problematic from an RL research perspective, but also from a safety perspective, as ignoring the delayed and prolonged effects of a drug can lead to drug overdosing related toxicity~\cite{guengerich2010}.

% Although the delayed and prolonged drug effect is a known phenomenon in medical sciences and pharmacodynamics, it is surprisingly missing from the applied RL literature handling personalized drug dosing. For example, none of the literature on using RL for closed loop Artificial Pancreas to control glucose levels considers the delayed and prolonged effect. This is wrong from an RL research perspective as it violates the Markov assumption. More alarmingly, this is dangerous from the medical science perspective as ignoring the delay and prolonged-ness can lead to drug overdosing, which can be toxic and damaging to the point of being fatal.
%% Precision drug dosing is a long standing problem in medicine \cite{vogen2010, tucker2016}. 

\paragraph{Contributions.}
In this paper, we identify prolongedness and delay as fundamental roadblocks to using RL in precision drug dosing, and focus on addressing the former. To that end, we introduce the \emph{prolonged action effect partially observable Markov decision process} (PAE-POMDP), a framework for modeling delayed action effects in decision making. We then present assumptions inspired by the pharmacology literature to convert the \emph{non-Markovian prolonged action effect} problem into a Markov decision process (MDP), therefore enabling the use of recent advances in model-free RL. To the best of our knowledge, our work is the \emph{first} to explore the prolonged effect of actions. We validate the proposed approach on a toy task, where the only violation of the Markov property comes from the prolonged effect of actions, and show that restoring the Markov assumption allows the RL agent we develop to significantly outperform previous baselines. We also address the challenging task of glucose control by optimal insulin dosing for type 1 diabetes patients
% , since insulin is a drug with low therapeutic index, 
by leveraging the open-sourced version of the FDA-approved UDA/Padova simulator~\cite{man2009, xie2018}. Although the glucose control task has garnered interest in the RL literature in the past~\cite{tejedor2020}, there does not appear to be a widely adopted reward function. Therefore, we design a clinically motivated reward function that explicitly avoids overdosing while effectively controlling glucose levels. Our results show that our approach of converting the PAE-POMDP into a MDP is not only competitive in terms of performance, but is also remarkably more time and memory efficient than the baselines, making it more suitable for real-time dosing control systems. %{\color{red} something about results. + discuss solution is compute and memory efficient}%
With these contributions, we aim to raise awareness of both the delayed and prolonged effects of drugs while tackling personalized drug dosing with RL, and hope that the proposed reward function will help break the entry barrier to control glucose levels for a closed-loop artificial pancreas, while fostering future research in this direction.\looseness-1

\paragraph{What this paper does not do.}
This paper does not aim to provide a general solution to tackle the prolonged effect of actions, but one crafted specifically for precision drug dosing that allows to quickly bring the problem back to the MDP framework in a compute- and memory-efficient way. % To further clarify, our solution does not handle qualitative action choices such as left, right etc. 
We do not provide a new algorithm to tackle prolongedness, but instead propose a pharmacologically motivated effective way to enable the use of  already existing RL algorithms for precision drug dosing. We are assuming drug action in isolation, instead of in combination with other drugs. Learning and predicting drug synergies is a growing body of research that can provide interesting future work.\looseness-1

\section{Related Work}
Blood glucose control in individuals with Type 1 diabetes is a longstanding problem in medicine that has stimulated the interest of RL researchers for a long time. In a systematic review of reinforcement learning approaches for blood glucose management, \citeauthor{tejedor2020}~(\citeyear{tejedor2020}) reported 347 papers between 1990 and 2019 on the topic, of which 11 used the UVA/Padova simulator \cite{man2009, man2014, xie2018}. While most of these work explore RL algorithms -- \eg Actor Critic, Q-learning, SARSA, and DQN among others -- to address the glucose control problem, none of them acknowledges the fundamental challenge of prolongedness and most consider the current blood glucose level as a sufficient statistic for state information. Even in the cases where actions are considered as part of the state information, only the most recent action is taken into account, which is not enough to restore the Markov assumption violated due to prolongedness (see appendix). \citeauthor{fox2020}~(\citeyear{fox2020}) formalized the problem as a POMDP by augmenting 4 hours of blood glucose and insulin information, but they did not recognize prolongeness of drug effects as the reason for partial observability. In fact, to handle prolongedness, the most recent blood glucose measurement and residual active insulin are sufficient, as we show in the experiments section.\looseness-1
% For the glucose control task, since the target area of action is bloodstream, delay can be small to negligible and hence neglecting it might not have a huge impact. But prolongedness will last longer and omitting it has serious drawbacks.

There is a similar theme of exploring various state of the art deep RL algorithms for other drug dosing problems. However, these works also miss the crucial point of recognizing and handling the prolonged effect of drugs. In a non-exhaustive list, \citeauthor{Nemati2016}~(\citeyear{Nemati2016}) use a Q-Network to learn a personalized heparin dosing policy, \citeauthor{Weng2017}~(\citeyear{Weng2017}) use policy iteration to manage glycemia for septic patients in critical care, \citeauthor{Lin2018}~(\citeyear{Lin2018}) propose the use of Deep Deterministic Policy Gradient (DDPG, \cite{lillicarp2016}) for heparin dosing in critical care patients, and \citeauthor{martinez2019}~(\citeyear{martinez2019}) use Double Deep Q-Networks (DDQN) to administer opioids for pain management in critical care patients. But none of these papers discusses the prolonged effect of drug doses. \citeauthor{zadeh2022}~(\citeyear{zadeh2022}) use Deep Q-Networks (DQN) to administer the anticoagulant warfarin, and they consider a pre-defined duration along with a dose. They use a Pharmacokinetic/Pharmacodynamic (PK/PD) model to determine the duration and add it to the state information, along with the patient information, blood coagulability measure and dosing history. However, the authors do not explicitly recognize prolonged drug effect as one of the reasons for adding the duration information. In addition, since dose response is individual specific, finding the right duration is a challenge in itself.\looseness-1

\section{Background}
\paragraph{Markov Decision Process (MDP).}
Reinforcement Learning (RL) is a framework for solving sequential decision making problems where an agent interacts with the environment and receives feedback in the form of reward. The typical formal framework for RL is the Markov Decision Process (MDP). A MDP $\mathcal{M}$ is a 5-tuple $(\mathcal{S}, \mathcal{A}, r, \mathcal{P}, \gamma)$, where $\mathcal{S}$ is a (finite) set of states, $\mathcal{A}$ is a (finite) set of actions, $\mathcal{P}$ is the state transition probability  $\mathcal{P}(s_{t+1}=s'|s_t=s, a_t=a)$, $r : \mathcal{S} \times \mathcal{A} \times \mathcal{S} \rightarrow \mathbb{R} $ is the reward function and  $\gamma \in [0, 1)$  is the discount factor. As the name suggests, a MDP obeys the Markov assumption that the future is in independent of the past given the present, which means that transitions and rewards depend only on the current state and action and not on the past history. The goal of an RL agent is to find a policy $\pi: \mathcal{S} \times \mathcal{A} \rightarrow [0, 1]$ that maximizes the cumulative discounted return $\sum_{t=0}^\infty \gamma^{t} r_t$.\looseness-1

\paragraph{POMDP and Recurrent Policy.}
A Partially Observable MDP (POMDP) is a 7-tuple $(\mathcal{S}, \mathcal{A}, r, \mathcal{P}, \Omega, \mathcal{O}, \gamma)$, where $\Omega$ is the set of observations, $\mathcal{O}$ is the set of conditional observation probabilities, $O(\omega|s)$, also known as the emission function, and the rest of the elements are the same as in an MDP. In a POMDP, the agent does not have direct access to the identity of the states, instead needing to infer them through the observations. Note that while in a MDP, an agent which aims to act optimally with respect to the expected long-term return only needs to consider Markovian policies, $\pi : \mathcal{S} \times \mathcal{A} \rightarrow [0, 1]$, in a POMDP, policies need to either rely on the entire history of action and observations, or to infer the distribution of  hidden states from this history (as in belief-based POMDP dolution methods). In recent work, recurrent networks have become the standard for implementing POMDP policies. 
We call recurrent policy a mapping $\pi : \Tau \times \mathcal{A} \rightarrow [0, 1]$   where $\Tau$ is the space of trajectories $\tau = \{(o_t, a_t, r_t)\}_{t=0}^T$ of up to $T$ time steps.\looseness-1

\paragraph{Q-Learning and Deep Q-Networks (DQN).}
Q-Learning \cite{watkins1989} is a  model-free RL algorithm that estimates $\mathcal{Q}: \mathcal{S} \times \mathcal{A} \rightarrow \mathbb{R}$, a value function assessing the quality of action $a_t \in \mathcal{A}$ at state $s_t \in \mathcal{S}$. The value function can be estimated as: $Q(s_t, a_t) = Q(s_t, a_t) + \alpha\big(r_t +\gamma \max _{a} Q(s_{t+1}, a) - Q(s_t, a_t)\big)$, %Equation \ref{eqn:q_learning}
where $\alpha \in (0, 1)$ is the learning rate. In DQN \cite{mnih2015}, the $Q$-values are estimated by a neural network  minimizing the Mean Square Bellman Error:   $L_i(\theta_i) = \mathbb{E}_{s_t, a_t \sim \pi_b} \big[(y_i - Q(s_t, a_t ; \theta_i))^2\big]   \\
    y_i = (r_t + \gamma \max_{a_{t+1}} Q(s_{t+1}, a_{t+1}; \theta_{i-1})$, where $\pi_b$ is the behavior policy, $y_i$ is the target and $\theta_i$ are the parameters of the $Q$-network at iteration $i$.\looseness-1

\iffalse
\begin{equation}
\label{eqn:q_learning}
    Q(s_t, a_t) = Q(s_t, a_t) + \alpha\big(r_t +\gamma \max _{a} Q(s_{t+1}, a) - Q(s_t, a_t)\big)
\end{equation}

\begin{equation}
\label{eqn:msbe}
\begin{split}
    L_i(\theta_i) = \mathbb{E}_{s_t, a_t \sim \pi_b} \big[(y_i - Q(s_t, a_t ; \theta_i))^2\big]   \\
    y_i = (r_t + \gamma \max_{a_{t+1}} Q(s_{t+1}, a_{t+1}; \theta_{i-1})
\end{split}
\end{equation}
\fi
% \paragraph{ADRQN}

\section{Method}

\subsection{Problem Formulation}
Drug dosing can be formalized as a POMDP, because the effect of a medication is felt over a period of time after its administration. In this section, we introduce \emph{prolonged action effect} POMDPs, referred to as PAE-POMDPs, a subclass of POMDPs in which an action's effect lasts more than one time step. % PAE-POMDP encapsulates all the drug dosing problems. 
More precisely, the action $a_t$ taken at time step $t$ continues to affect the future states of the environment for $\kappa \in \mathbb{Z^+}$ time steps, where $\kappa$ is the time interval necessary for $a_t$'s effect to fall below a given threshold. The value of $\kappa$ is environment specific and also depends on the amplitude of the action $a_t$. The above-defined prolongedness offers a \emph{forward view} in time. From a \emph{backward view} perspective, the state at a time step $t$ is the result of super-imposing the effects of the actions from several preceding time steps, which are still felt at time $t$ i.e. $s_{t+1} = s_t + \sum_{k=0}^{\kappa} \Delta_{a_{t-k}}$, where $\Delta_{a_{t}}$ is the effect of action $a_t$ on the state. In other words, the Markov transition function $P(s_{t+1}|s_t, a_t)$ is no longer valid, and instead the transition function becomes $P(s_{t+1}|s_t, (a_{t-k})_{k=0}^\kappa)$. In this case, the reward function can still be assumed to be Markov $R(s_t, a_t, s_{t+1})$ (see Equation \ref{eqn:reward}). From the perspective of the agent, this means that it needs to keep track of the history of actions over a preceding period of time. In general, this problem is no simpler than a regular POMDP, as an agent that keeps track of its history may need to remember all the actions taken since the beginning of time. However, in this work, we consider a more circumscribed problem formulation which is relevant for therapeutic dosing. 
Specifically, we use knowledge from pharmacology to enable us to restore the Markov assumptions for drug-dosing PAE-POMDP.\looseness-1

\subsection{Converting PAE-POMDP into MDP}
\label{sec:approach}
In this section, we present the specific knowledge used to restore the Markov assumptions in drug dosing problems and show how by leveraging them we can effectively convert a PAE-POMDP into an MDP.\looseness-1
% In this section we enlist the assumptions used in converting PAE-POMDP into MDP and comment on why these assumptions are practical in the context of precision drug dosing:\\
%\begin{itemize}
    %\item 
 \paragraph{Action and effect equivalence.} We start by noting that as per the pharmacodynamics axioms, drug effects are determined by drug concentration at the site of action~\cite{holford1984, nick2018}. Therefore, action and effect are considered to be equivalent and used interchangeably hereinafter.\looseness-1 %So, the decay in the action amplitude (drug concentration) is considered to be equivalent to the decay in its effect.

   \paragraph{Exponential rate of decay.} % Biological substances such as drugs have biological half-life, which is the time required by a biological system to eliminate half of the substance from its initial concentration by natural process~\citep{}. 
    Motivated by the pharmacology~\cite{benet1995, dasgupta2020}, biotechnology~\cite{hobbie2007}, as well as chemical- kinetics~\citep{atkins2006} literature, we adopt an exponential decay  model of drug concentration over time~\cite{annamalai2010}. 
    % The exponential decay model has previously been used in pharmacology \cite{dasgupta2020} and biotechnology literature to model drug concentration over time \cite{annamalai2010}. The same model also appears in the rate laws in chemical kinetics \citep{atkins2006}.\looseness-1
    % The rate law is a formal expression defining the relationship between the forward rate of a chemical reaction and the initial concentration of the compounds. The integrated rate law for a first order reaction depending on the concentration of a single reactant is defined as: 
    % \[ln[A] = -kt + ln[A]_0\] where $[A]_0$ is the initial concentration, $[A]$ is the concentration at time $t$ and $k$ is the rate constant. This type of law governs for example insulin concentration which can be used to control diabetes~\citep{ferr1979}.  If we consider the initial concentration as an action, this equation can be viewed as describing the evolution of this action's effect over time. By exponentiating both sides, we can see that the future effect of an action is proportional to its initial magnitude, and it decays exponentially over time. Hence, we will define a PAE-POMDP using these modelling assumptions.
    In particular, we assume that the initial action effect will decay at a constant rate $\lambda$, which is specific to the environment (drug as well as individual). Formally, we assume that $a_{>t \leq \kappa} = \lambda a_t$. Note that  here we implicitly assume that the action's effects can safely be ignored after $\kappa$ time-steps. This assumption is valid in the context of drug dosing, since drugs are ineffective below a certain concentration. The observation of the agent at a particular time step can therefore be defined conditional on the current state and on all the past actions: $o_{t+1} \sim \mathcal{O}(.|s_t, a_{<t})$. Note that although this is a generalized notation, not all the actions $a_{<t}$ that occur before $t$ have an effect on $o_{t+1}$ (only those within a $\kappa$ window). If the amplitude of an action decays by a factor of $\lambda$ for each of the $\kappa$ time steps during which the action's effect is felt, then at any time-step $t$, the environment state is a function of a sub-sequence of the previous actions $\{a_{\leq t}\}$, which we call \emph{effective action}, $a_{t_\mathcal{E}}$, and define as:
\begin{equation}
\label{eq:effective_action}
a_{t_\mathcal{E}} =
\begin{cases}
    0, &\text{if $t=0$}\\
    \lambda a_{(t-1)_\mathcal{E}} + a_{t-1}, &\text{otherwise}\\
\end{cases}
\end{equation}
where $\lambda \in (0, 1)$ enforces the exponential decay assumption on the actions. Note that setting $\lambda = 0$ falls back to the classical RL scenario, whereas setting $\lambda = 1$ entails a infinitely long action effects.\looseness-1

    \paragraph{Additive composition of action effects.} We note that Equation~\ref{eq:effective_action} leverages the
    additive property of drug concentration, and therefore assumes that action effects are additive and independent of each other. 
    % , which means that we do not need to model interactions. 
    With the additive action effects assumption, the conditional distribution $\mathcal{O}(.|s_t, a_{<t})$ could be modelled as a sum of effects due to the actions.\looseness-1
    % One further specification relates to the way in which action effects compose with each other. Drug concentration having the additive property, in the precision dosing context we can assume that action effects are additive and independent of each other, which means that we do not need to model interactions. Note that with the additive action effects assumption, the conditional distribution $\mathcal{O}(.|s_t, a_{<t})$ could be modelled as a sum of effects due to the actions.

    %\item 

    %\item 

\paragraph{From PAE-POMDP to MDP.} Finally, we convert the PAE-POMDP into a MDP, by augmenting the states with effective actions. Formally, we define a revised MDP, $\mathcal{M_E} = (\mathcal{S_E}, \mathcal{A}, r, \mathcal{P}, \gamma)$, where the states in  $\mathcal{S_E}$ are defined as $s_{t_\mathcal{E}} = (s_t, a_{t_\mathcal{E}})$. The updated state definition restores the Markov assumption, and therefore $\mathcal{M_E}$ can be solved as a proxy of the original MDP $\mathcal{M}$. Note that $\mathcal{M_E}$ and $\mathcal{M}$ only differ in  their state space. Since we have access to $\mathcal{M_E}$, we can  use traditional RL algorithms to solve it.\looseness-1
%\end{itemize}

% Specifically, we are going to use a decay assumption inspired from 
% This kind of natural phenomenon is common in natural dynamical systems (e.g. heat dissipation or radioactive decay).

% We assume that the action effects are additive, which is relevant in the drug dosing context since a new drug dose increases the plasma concentration of the drug by the magnitude of the dose \cite{nick2018}

% More precisely, suppose the action $a_t$ taken at time step $t$ continues to change the future states of the environment for $\kappa \in \mathbb{Z^+}$ time steps. The value of $\kappa$ is environment specific and also depends on the amplitude of the action $a_t$. %The change in the states due to $a_t$ is $\Delta a_t$.%
%In this paper, we assume this is the only source of partial observability. The effective length $\kappa_A$ of action $a_t$'s effect depends on its amplitude, i.e. $\kappa_{a_t} \propto |a_t|$.

% Moreover, note that if action effects are additive, this conditional probability distribution could be modelled as a sum of effects due to the actions. This assumption is less important but leads to further computational gains.

% \subsection{Transforming PAE-POMDP into an MDP}

\section{Experiments}
\begin{figure}
\centering
  \includegraphics[width=0.7\linewidth]{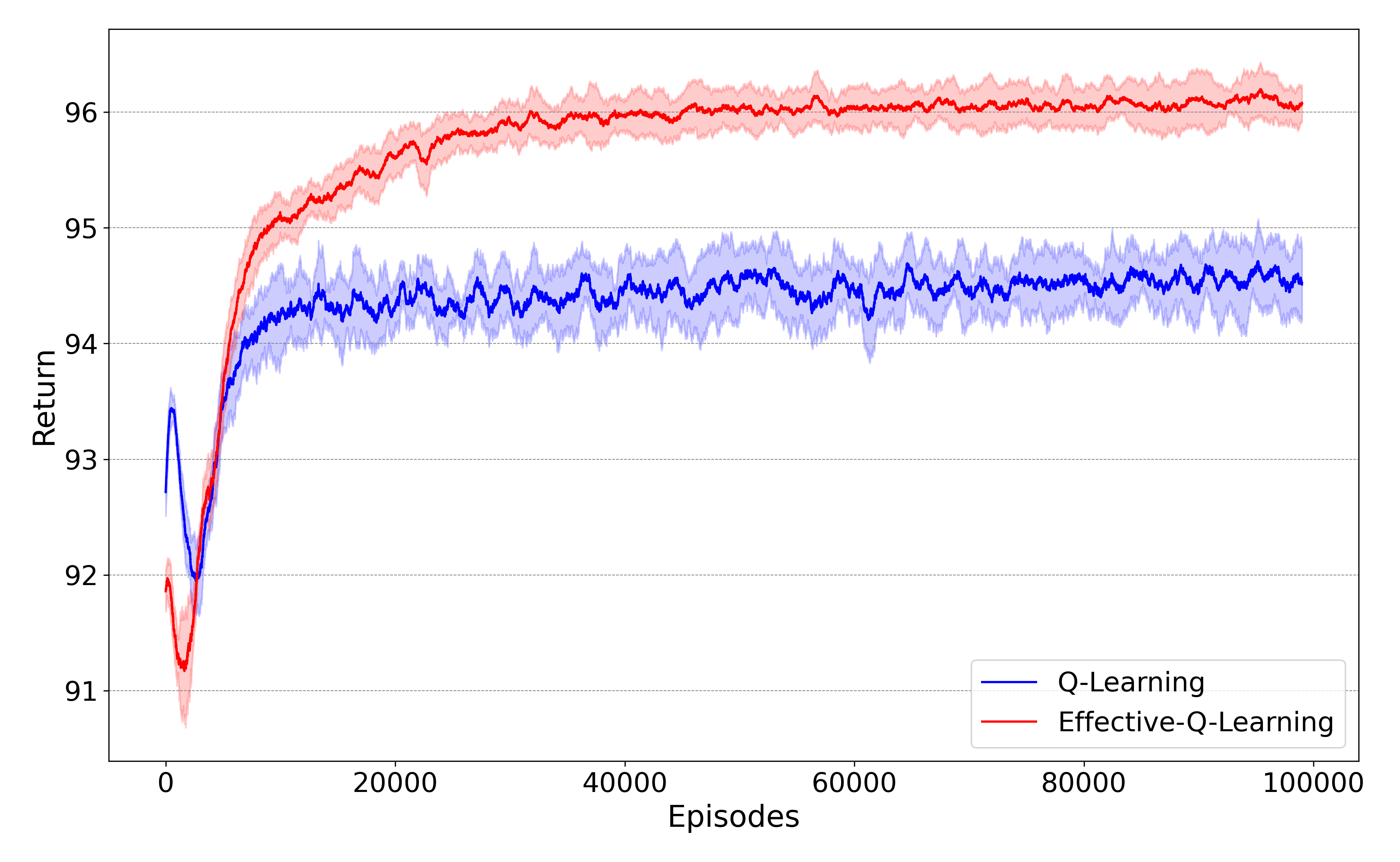}
\caption{\textbf{Performance on MoveBlock.} Cumulative discounted return of tabular Q-learning and Effective Q-learning. Mean and std of performance over 10 seeds. Performance is smoothed with a window of size 1000.\vspace{-0.5cm}}%\looseness-1}
\label{fig:tabular}
\end{figure}

We validate the proposed approach on a toy environment, as a sanity check to prototype our method, and on a challenging glucose control environment. This section describes both environments and presents the obtained results. %, by contrasting them with baselines.

\subsection{Toy Environment: \emph{MoveBlock}}
\paragraph{Environment.} We created the MoveBlock environment to assess our proposed approach in a controlled manner. In particular, we built the environment in such a way that the only violation of the Markov property comes from the prolonged effect of actions. In MoveBlock, the task is to move a block from its random initial position to a fixed final position on a slippery surface with minimum possible force applied along the horizontal axis in the direction of the goal. Due to the slippery nature of the floor, an exerted force will cause the object to move until it is stopped by friction, prolonging the effect of actions over multiple time-steps. The reward system is the same as the one in the Mountain Car environment \citep{brockman2016openai} -- \ie there is a high reward for reaching the goal position, but action magnitude proportional penalties to encourage policies where minimum effort is used. The observation is the position (continuous) while the velocity remains unobserved to the agent.

\paragraph{Experimental setup and results.} To keep the states tractable, we discretized the continuous position values into 10k states. The action space is discrete, but to keep the effective actions tractable, we clipped the effective action value to the maximum action value. We then trained a tabular Q-learning agent, which serves as baseline, with $\epsilon$-greedy exploration. Next, we modified this tabular Q-learning agent with the method proposed in Section \ref{sec:approach}, and trained an \emph{Effective Q-learning} agent. The results are shown in Figure~\ref{fig:tabular}. We observe that the Effective Q-learning agent, which restores the Markov assumptions broken by the prolonged action effects, consistently outperforms the Q-learning agent despite using the same Q-learning algorithm, which highlights the importance of considering the prolongedness of actions when solving PAE-POMDPs.

\subsection{Glucose Control Environment}

\paragraph{Environment.} We chose blood glucose control as a real-world example of PAE-POMDP given the low therapeutic index of insulin, which makes it a good candidate for precision dosing. We used Simglucose \citep{xie2018}, the OpenAI Gym \citep{brockman2016openai} implementation of the FDA approved UVA/Padova Type1 Diabetes Mellitus Simulator (T1DMS)~\citep{man2009, man2014} built on an in-silico population of 30 patients. In this case, the task is to manage blood glucose by administering insulin to type 1 diabetes patients, who lose their ability to produce insulin for the rest of their lives. In particular, the goal is to maintain blood glucose as close as possible to the non-diabetic range for a given patient. However, the environment does not come with a recognized pre-defined reward function.

\paragraph{Experimental setup.} We are interested in training a personalized glucose control policy instead of a population-level policy. To this end, we randomly selected a virtual patient from the simulator and modified the simulator for the experiments presented in this paper (See Appendix for more details). In this PAE-POMDP setting, the states are hidden patient parameters~\cite{kovatchev2009}, observations are continuous blood glucose readings, actions are discrete insulin doses in the range of $[0, 5]$, transition dynamics are specific to the chosen patient but unknown to the agent, and the reward function is described in the next paragraph. We appended the current observations with the effective actions as described in section \ref{sec:approach}. 

\paragraph{Reward.}
Reward design in RL is a challenging task in itself. The task of maintaining blood glucose as close as possible to the non-diabetic range, can be addressed by maximizing the time spent in the target zone with as little insulin as possible. The minimal insulin dose requirement is due to the need to avoid building insulin resistance, a condition where a person no longer responds to small doses of insulin.
% and needs more insulin to absorb the same amount of blood glucose as before.
Note that the tendency to develop resistance is applicable to any drug. Unfortunately, recently introduced reward functions do not fulfill the above-discussed criteria. Therefore, we designed a biologically inspired custom \textit{zone reward} that incentivizes the time spent in the target zone and penalizes hyperglycemia, hypoglycemia and high insulin doses. Formally, the reward is defined as:
\begin{equation}
\label{eqn:reward}
r_t(s_{t-1}, a_t, s_t) = r_{t, state}(s_{t-1}, s_t) - r_{t, action}(a_t)
\end{equation}
where $r_{t, state}$ is a state reward which encourages the agent to maintain the target vital statistics within a healthy range as follows:
\small
\begin{equation}
\label{eq:r_state}
\begin{split}
r&_{t, state}(s_{t-1}, s_t) = \\
&
\begin{cases}
    -100 & s_t<70 \text{ or } s_t>200 \text{ (episode termination)}\\
    -1 & s_t<100 \text{ and } s_t - s_{t-1} < 0.5 \text{ (hypoglycemia)}\\
    -1 & s_t>150 \text{ and } s_t - s_{t-1} > 0.5 \text{ (hyperglycemia)}\\
    10 & 100<=s_t<=150 \text{ (target blood glucose)}
\end{cases}
\end{split}
\end{equation}
\normalsize
and $r_{t, action}$ is a penalty term which encourages the agent to avoid overdosing:
\begin{equation}
\label{eq:a_state}
r_{t, action}(a_t) = 0.1 * a_t^2 \quad \text{(penalize overdosing)}
\end{equation}
Note that this kind of reward function is well-suited for any homeostatic control problem in drug dosing.

\begin{figure}
\centering
  \centering
  \includegraphics[width=0.7\linewidth]{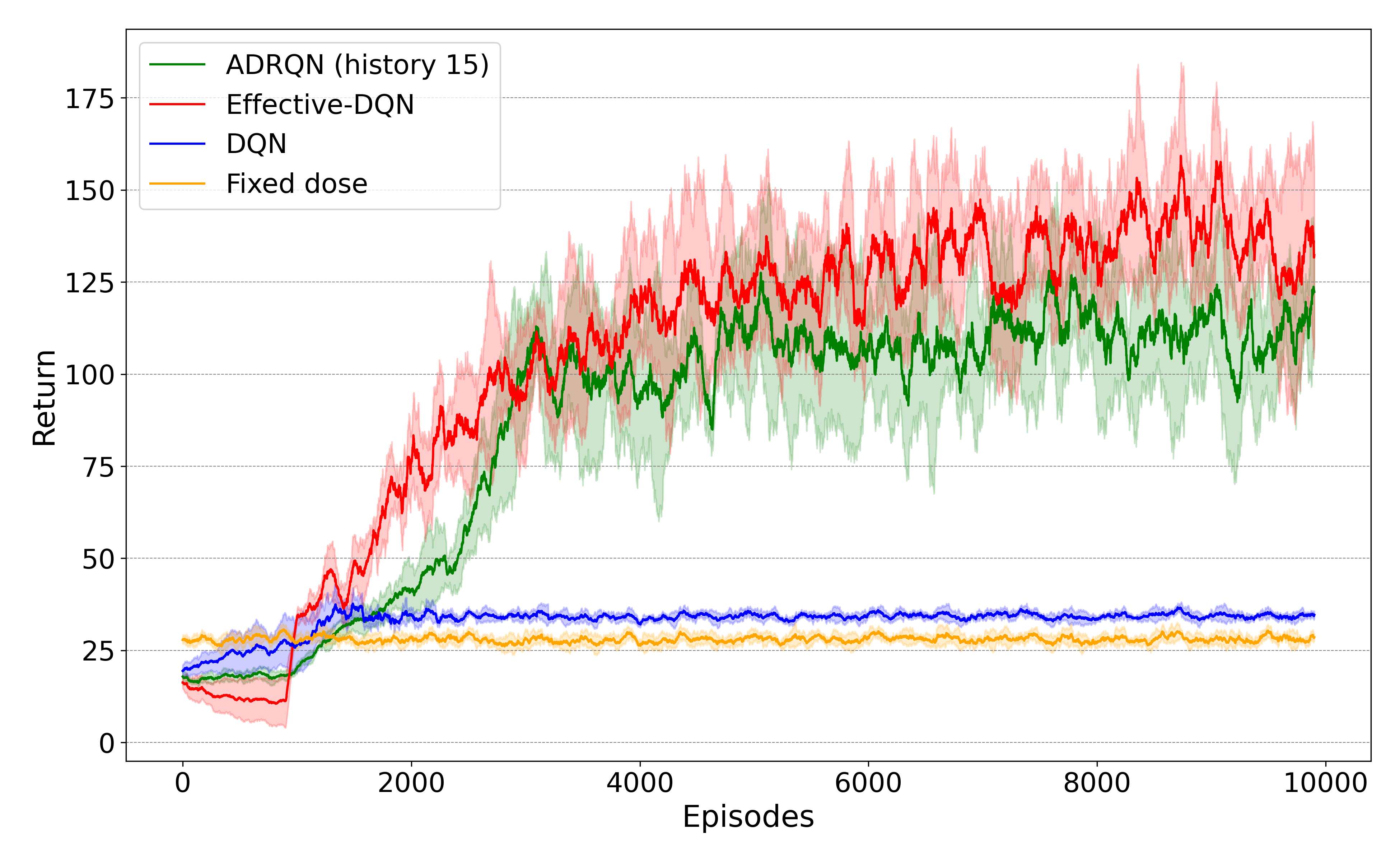}
\caption{\textbf{Performance (Cumulative discounted sum of rewards) on the glucose control task.} Since DQN assumes the environment to be Markov, its performance is almost as poor as the fixed policy. The performance of Effective-DQN is at par with ADRQN (history length 15); both consider prolonged effects of actions.\looseness-1}
\label{fig:diabetes_perf}
\end{figure}

\begin{figure}
\centering
\includegraphics[width=0.8\columnwidth]{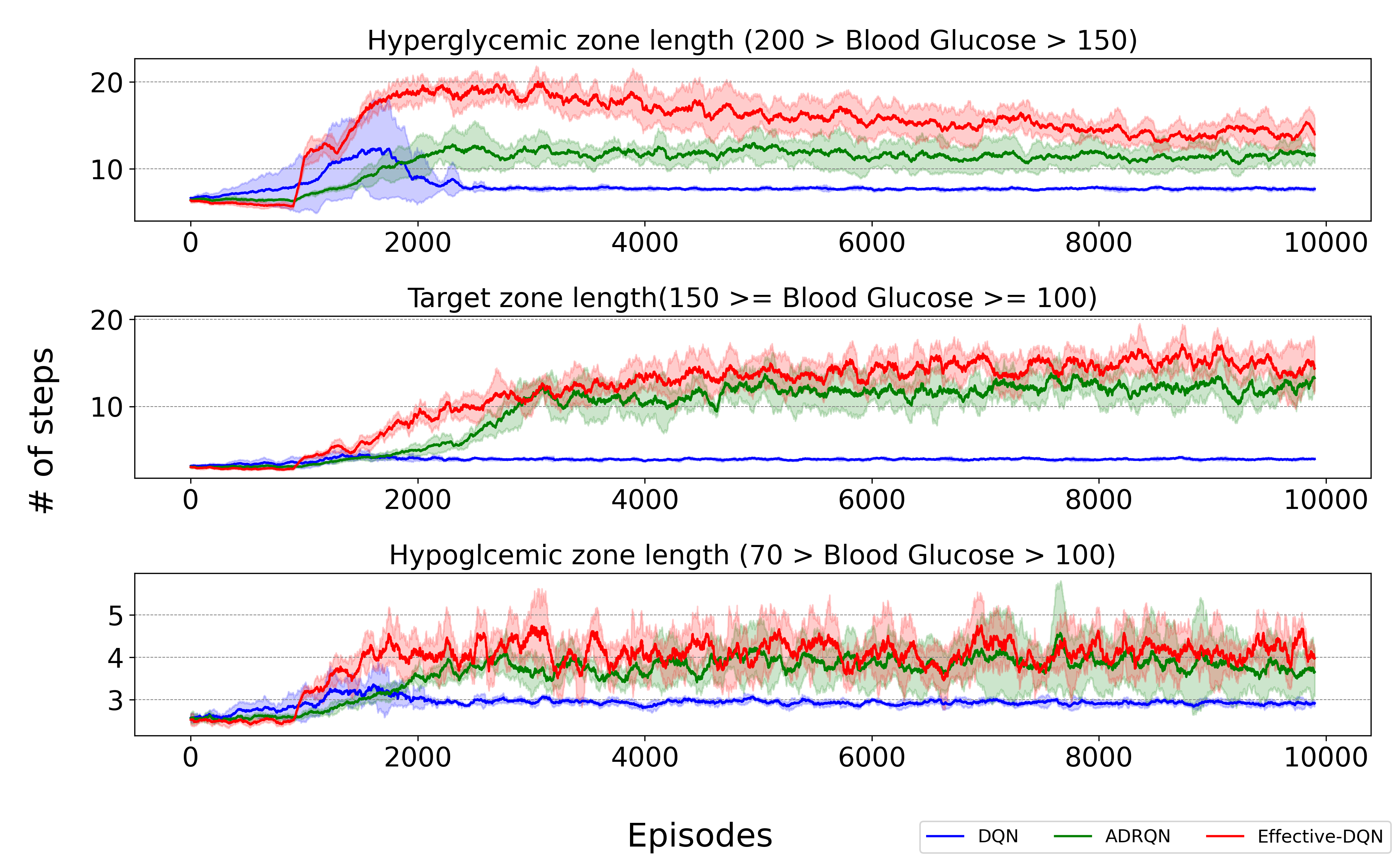}
\caption{\textbf{Glucose control performance breakdown.}  Average number of steps in each blood glucose zone over training episodes (std over 5 seeds). Due to the prolonged action effect, Effective-DQN spends more time in the hyperglycemic zone as well as in the target zone than ADRQN. DQN learns a sub-optimal policy that tends to overdose in order to to quickly get out of hyperglycemia and eventually leads to episode termination due to hypoglycemia.\looseness-1}
\label{fig:diabetes_zone_breakdown}
\end{figure}
\iffalse
\begin{table}[b]
  \caption{Runtime and memory requirements of 5 agents trained on a Nvidia RTX 8000 GPU with the same 5 seeds.}
  \label{table:compute-table}
  \centering
  \resizebox{\columnwidth}{!}{
  \begin{tabular}{lcc}
    \toprule
    Model     &  Runtime (mean $\pm$ std) & Max memory use \\
              & (mins/10K episode)                        & (MiB) \\
    \midrule
    % DQN & $58.11 \pm 3.94$ & 1363 \\
    ADRQN (history 15) & $359.64 \pm 19.25$ & 1623 \\
    \textbf{Eff-DQN (ours)} & $ \textbf{129.04} \pm \textbf{42.73} $ & \textbf{1513} \\
    \bottomrule
  \end{tabular}
  }
\end{table}
\fi

\begin{table*}
\begin{minipage}{0.33\linewidth}
\vspace{0.6cm}
  \centering
  \resizebox{\linewidth}{!}{
  \begin{tabular}{lcc}
    \toprule
    Model     &  Runtime & Max \\
              & %(mins/10K episode)                      
              & memory \\
    \midrule
    % DQN & $58.11 \pm 3.94$ & 1363 \\
    ADRQN %(history 15) 
    & $359.64 \pm 19.25$ & 1623 \\
    \textbf{Effective-DQN} & $ \textbf{129.04} \pm \textbf{42.73} $ & \textbf{1513} \\
    \bottomrule
  \end{tabular}}
  \captionof{table}{\textbf{Runtime and memory requirements.} 5 agents trained on a Nvidia RTX 8000 GPU with the same 5 seeds (mean $\pm$ std). Runtime is in mins/10K episode and memory in MiB.}
  \label{table:compute-table}
\end{minipage}\hfill
\begin{minipage}{0.64\linewidth}
\centering
\begin{subfigure}{.5\textwidth}
  \centering
  \includegraphics[width=0.93\linewidth]{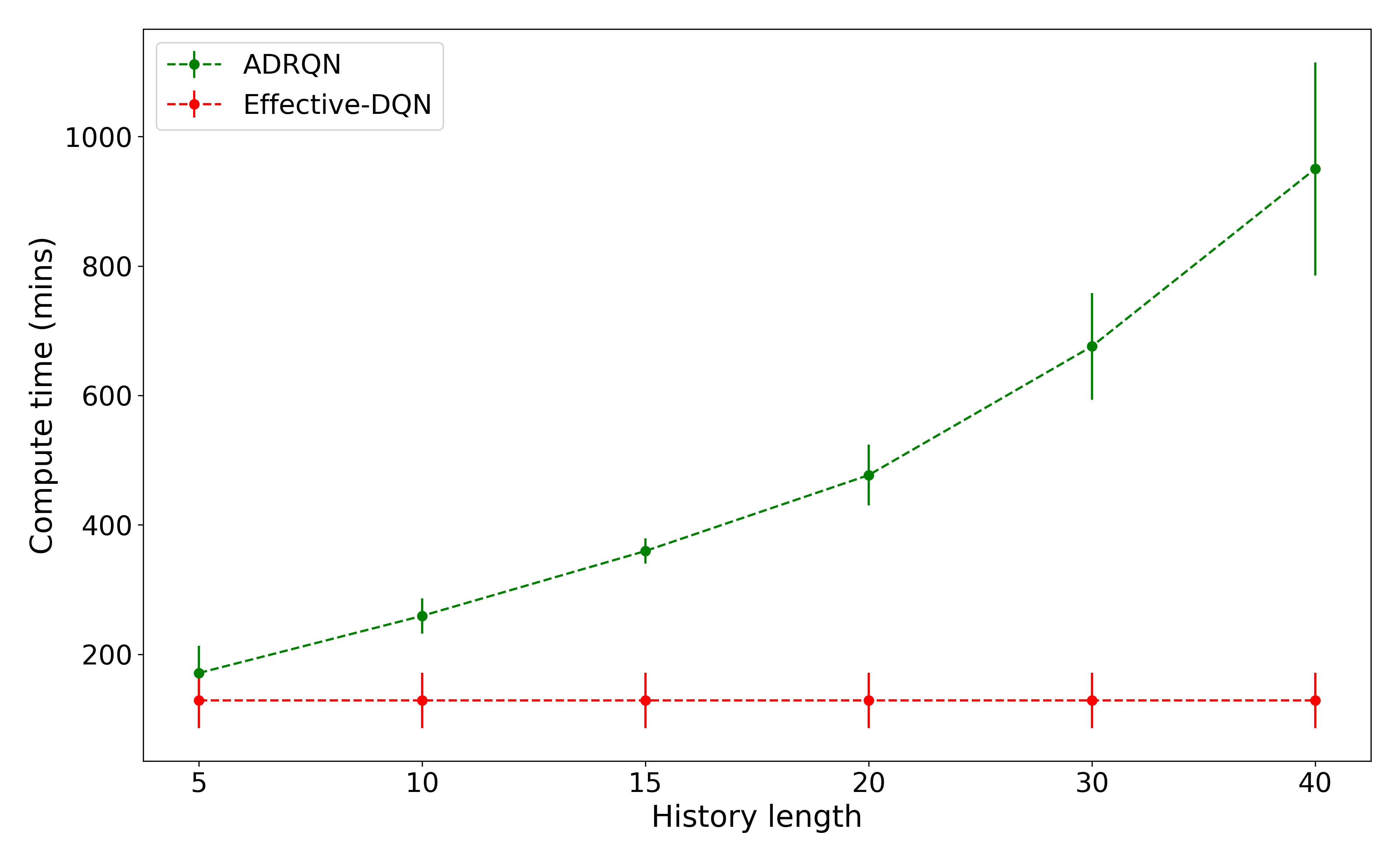}
  %\caption{Compute}
  \label{fig:ADRQN_compute}
\end{subfigure}%
\begin{subfigure}{.5\textwidth}
  \centering
  \includegraphics[width=0.93\linewidth]{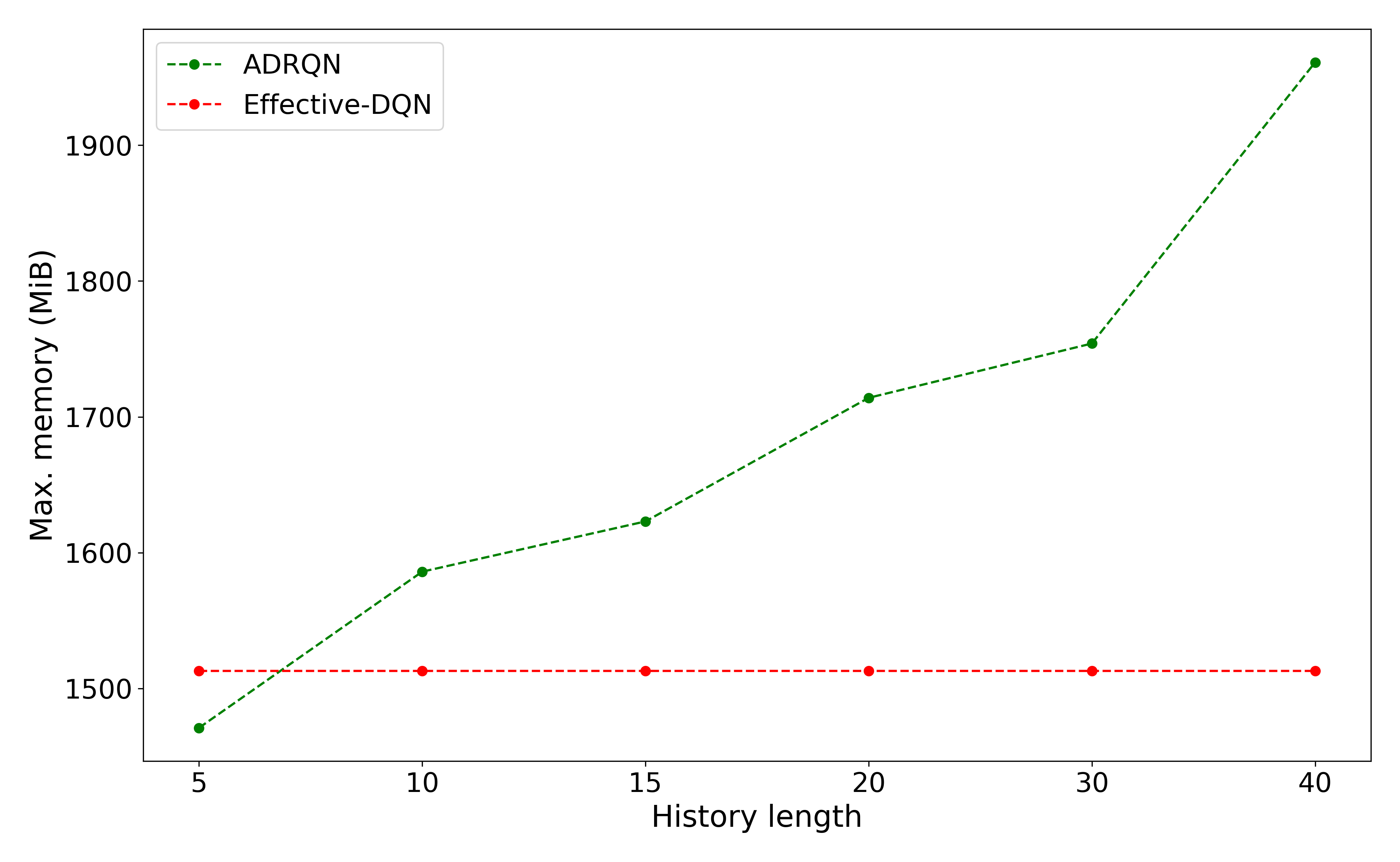}
  %\caption{Memory}
  \label{fig:ADRQN_memory}
\end{subfigure}
\captionof{figure}{\textbf{Effect of history length on compute (left) and memory requirement (right).} Compute time increases exponentially and memory requirement linearly with the history length for ADRQN agents, but Effective-DQN compute and memory are independent of the history length and hence remain  constant.\looseness-1}
\label{fig:ADRQN_compute_memory}
\end{minipage}
\end{table*}

% \begin{figure*}
% \centering
% \begin{subfigure}{.5\textwidth}
%   \centering
%   \includegraphics[width=\linewidth]{figs/performance_comparison_diabetes.png}
%   \caption{Performance}
%   \label{fig:diabetes_perf}
% \end{subfigure}%
% \begin{subfigure}{.5\textwidth}
%   \centering
%   \includegraphics[width=\linewidth]{figs/effect_of_history_len_diabetes_lr0.001.png}
%   \caption{Effect of History length in ADRQN}
%   \label{fig:ADRQN_history}
% \end{subfigure}
% \caption{\textbf{Glucose control results.} (a) Performance of Effective DQN is at par with ADRQN (history length 15). (b) ADRQN agent does not improve beyond a history length 10, but we chose a more stable agent with history length 15. Solid line is mean and shaded region is standard deviation over 5 seeds. All results are smoothed with a window of size 100.}
% \label{fig:diabetes}
% \end{figure*}

\subsubsection{Policies under comparison.}

All agents are trained using the \textit{zone reward} function introduced in the previous section.
\begin{enumerate}
    \item \textbf{Fixed Dose.} This is a handcrafted step-wise policy with the underlying assumption that higher blood glucose levels call for higher doses. Specifically, we inject the maximum insulin (5 units) for blood glucose in the range of 190 to 200. For every 10 units drop in blood glucose level from 190, we reduce the insulin dose by 1 unit until it is cut off at blood glucose 150. 
    \item \textbf{DQN.} 
    The DQN agent assumes the environment to be Markov, and hence is a good baseline to capture the effect of ignoring the inherent prolongedness of the actions.
    \item \textbf{Action-specific Deep Recurrent Q-Network (ADRQN)~\cite{zhu2017}.}  This baseline leverages a recurrent neural network that encodes the action history and appends it with the current observation to form a state. The action history length is set to 15, as per hyper-parameter tuning. Note that since we are the first to acknowledge the prolonged effect of drug dosing, we are also the first to propose an action history based baseline in this context. In a scenario where compute and memory is not a limiting factor, ADRQN (or any other action history based policy) can serve as the state of the art.\looseness-1
    \item \textbf{Effective-DQN.} This approach enhances the DQN baseline, but with the decay assumption introduced in the previous section i.e. the blood glucose values are augmented with the effective insulin dose to represent the state.
\end{enumerate}

\paragraph{Quantitative Results.}
% \begin{enumerate}
    %\item \textbf{Performance.} 
    We compare the cumulative discounted sum of rewards for all the agents under consideration in Figure \ref{fig:diabetes_perf}. The fixed dose policy, which undergoes no learning, achieves the lowest performance. Unsurprisingly, the DQN agent only slightly improves upon this naive hand-crafted policy, since it assumes the environment is Markov and thus ignores the prolonged effect of actions. The ADRQN agent exhibits a significant performance boost as compared to the DQN agent, since the learned policy depends not only on the current observation, but also on the action history responsible for the prolonged effect. The Effective-DQN agent's performance is on par with the ADRQN agent, suggesting that our decay assumption is a good approximation for the prolonged action effect observed in the diabetes simulator (see Appendix for details). To gain further insights, we stratify the performance of each RL agent into the hyperglycemic, hypoglycemic, and target zones, and assess the number of steps each agent spends in each blood glucose zone over training episodes (see Figure~\ref{fig:diabetes_zone_breakdown}). We observe that the DQN agent minimizes the negative reward for hyperglycemia and spends minimum time in the hyperglycemic zone. However, due to the prolonged effect of the actions taken in hyperglycemia, the agent spends very little time in the target zone as well as hypoglycemic zone, suggesting that the agent learns to overdose to quickly get out of hyperglycemia, and  eventually leading to episode termination due to hypoglycemia. Both ADRQN and Effective-DQN spend a similar amount of time in the hypoglycemic zone, whereas Effective-DQN spends slightly more time in hyperglycemia than ADRQN, increasing the time it spends in the target zone. This explains the performance gain of the Effective-DQN agent over ADRQN.\looseness-1
    
    %\item \textbf{Efficiency.}
 \paragraph{Efficiency analysis.} Since time sensitivity and limited memory are crucial aspects of real-time control systems, we compare the best performing agents (ADRQN and Effective-DQN) in terms of compute and memory efficiency. In Table~\ref{table:compute-table}, we present the compute and maximum memory requirement to train the models on the glucose control task. The \emph{Runtime} column presents the time required to train the models for 10K episodes on a single Nvidia RTX 8000 GPU. We used the readily available ADRQN PyTorch implementation\footnote{\texttt{https://mlpeschl.com/post/tiny\_adrqn/}}, and ensured both ADRQN and Effective-DQN are comparable in terms of capacity. Although some of the differences might still be attributed to the implementation, many are due to the individual properties of each model. In particular, ADRQN has the longest runtime as it needs to learn the action history through the LSTM~\cite{hochreiter1997} unit, and as shown in Figure~\ref{fig:ADRQN_compute_memory} (left), the compute time grows exponentially with the history length. History length depends on how prolonged the action effects are, which in turn are individual and drug specific factors. However, the compute time for Effective-DQN remains constant no matter the history length, making the compute efficiency gap between ADRQN and Effective-DQN increase for longer history lengths.
    In the \emph{max memory} column of Table~\ref{table:compute-table}, we present the maximum memory required by the models during training. We keep a replay buffer of the same size to train all the models. As shown in the table, ADRQN requires slightly more memory than Effective-DQN. While this difference does not appear significant, it is worth noting that ADRQN requires memory  proportional to the history length (see Figure~\ref{fig:ADRQN_compute_memory} (right)). Moreover, in order to benefit from longer histories, ADRQN might also require larger models, further increasing its memory requirement. In contrast, the Effective-DQN's memory requirement remains constant as we increase the history length.\looseness-1

\paragraph{Qualitative results.}

% \begin{figure*}[t!]
% \centering
% \begin{subfigure}{.49\textwidth}
%   \centering
%   \includegraphics[width=0.99\linewidth]{figs/evaluate_effdqn.png}
%   \caption{Effective-DQN}
% %   \label{fig:eval_effdqn}
% \end{subfigure}%
% \begin{subfigure}{.49\textwidth}
%   \centering
%   \includegraphics[width=0.99\linewidth]{figs/evaluate_adrqn.png}
%   \caption{ADRQN}
% %   \label{fig:eval_adrqn}
% \end{subfigure}
% \begin{subfigure}{.49\textwidth}
%   \centering
%   \includegraphics[width=0.99\linewidth]{figs/evaluate_dqn.png}
%   \caption{DQN}
% %   \label{fig:eval_dqn}
% \end{subfigure}
% \begin{subfigure}{.49\textwidth}
%   \centering
%   \includegraphics[width=0.99\linewidth]{figs/evaluate_fixed.png}
%   \caption{Fixed Dose}
% %   \label{fig:eval_fixed}
% \end{subfigure}
% \caption{\textbf{Policy visualization.} Trained greedy agents evaluated on 10 episodes with the same starting blood glucose. Red cross marks the end of episodes. The insulin administered is as prescribed by the respective policy. In the blood glucose plot, the space between the two red dotted lines is the target zone. Effective-DQN appears to be conservative and thus does not allow the patient to get into hypoglycemia. ADRQN agent gets into both hypo- and hyperglycemia.}
% \label{fig:evaluate}
% \end{figure*}

\begin{figure*}[t!]
\centering
\begin{subfigure}{.37\textwidth}
  \centering
  \includegraphics[width=0.95\linewidth]{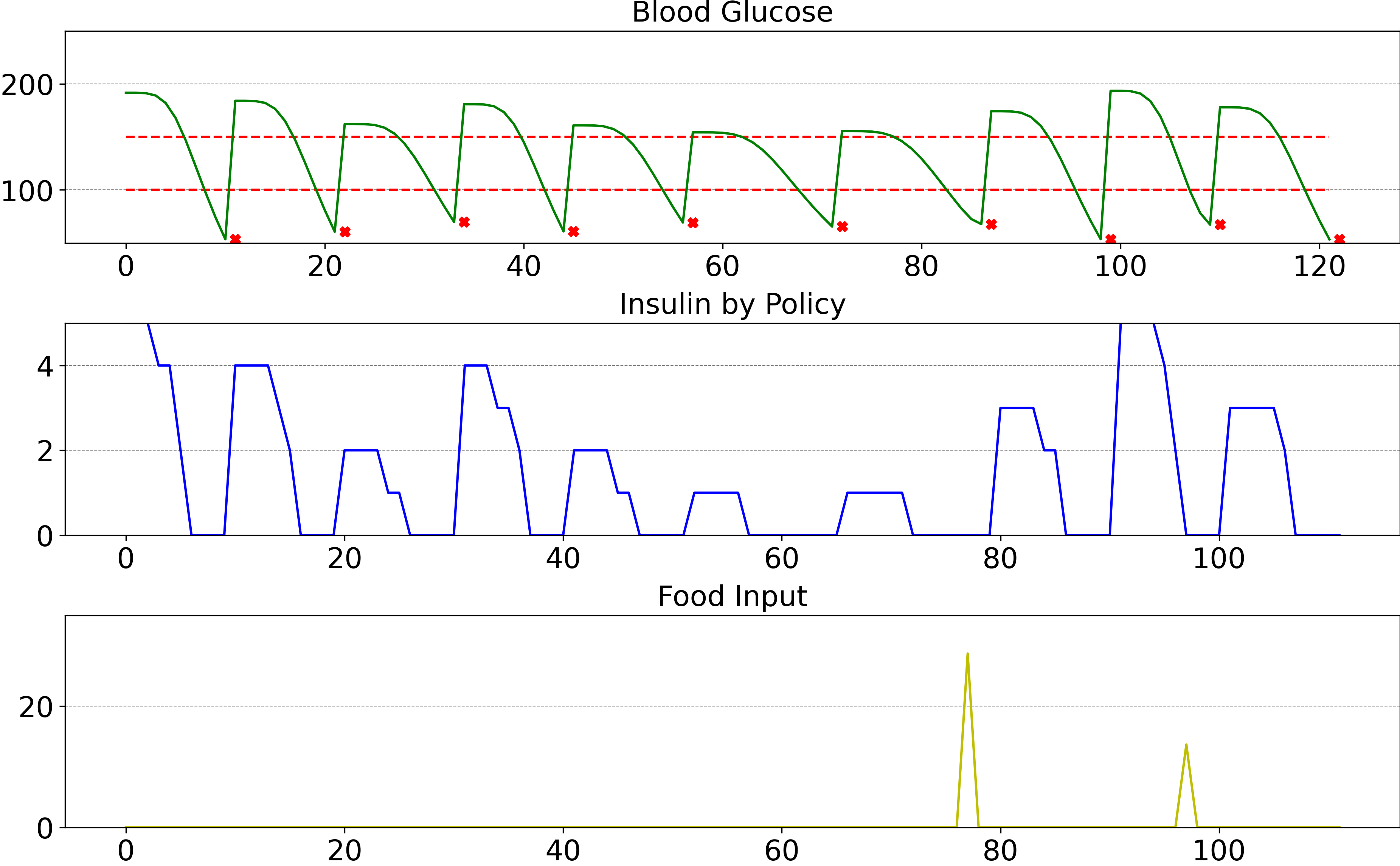}
  \caption{Fixed Dose}
%   \label{fig:eval_fixed}
\end{subfigure}%
\begin{subfigure}{.37\textwidth}
  \centering
  \includegraphics[width=0.95\linewidth]{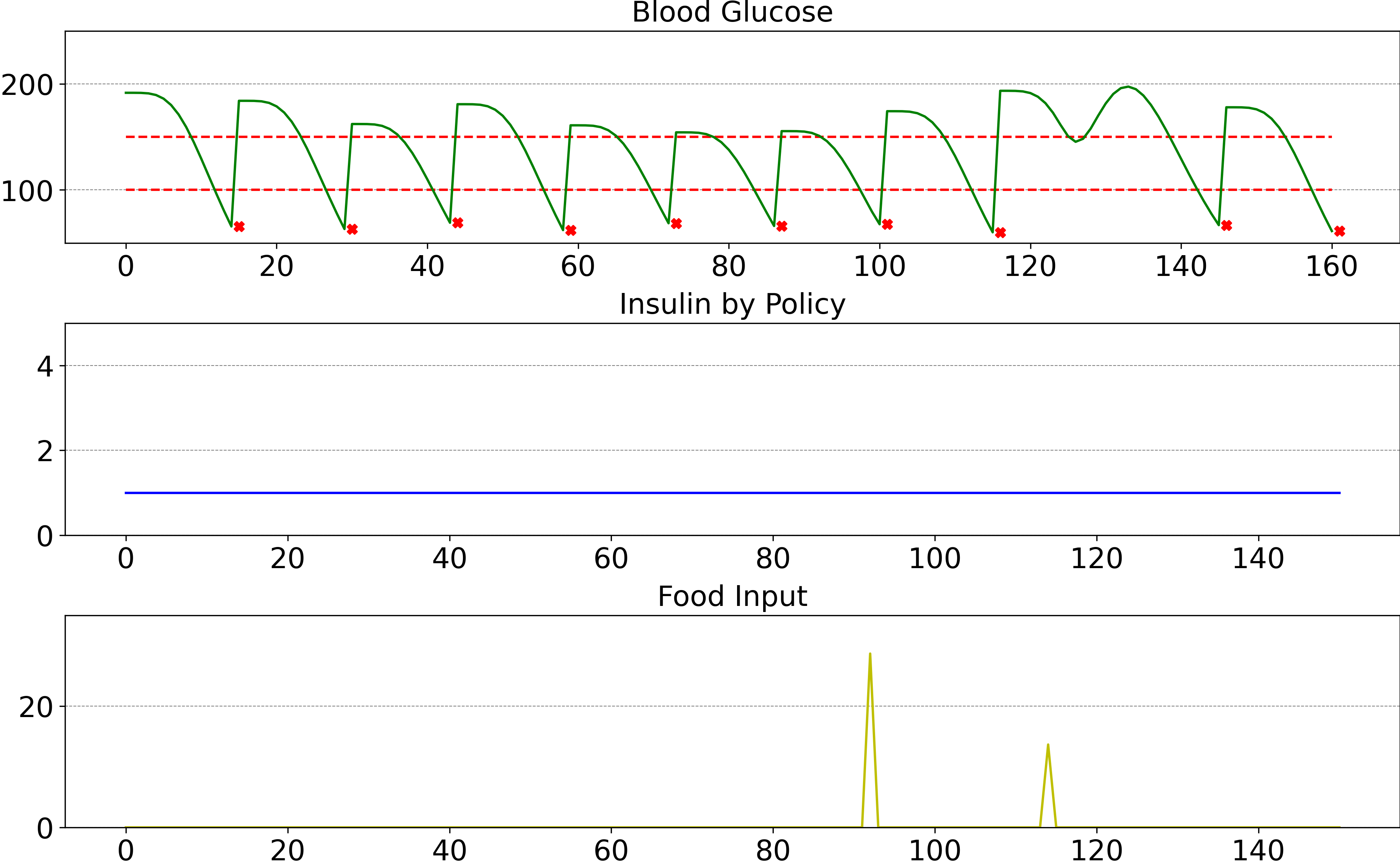}
  \caption{DQN}
%   \label{fig:eval_dqn}
\end{subfigure}
\begin{subfigure}{.37\textwidth}
  \centering
  \includegraphics[width=0.95\linewidth]{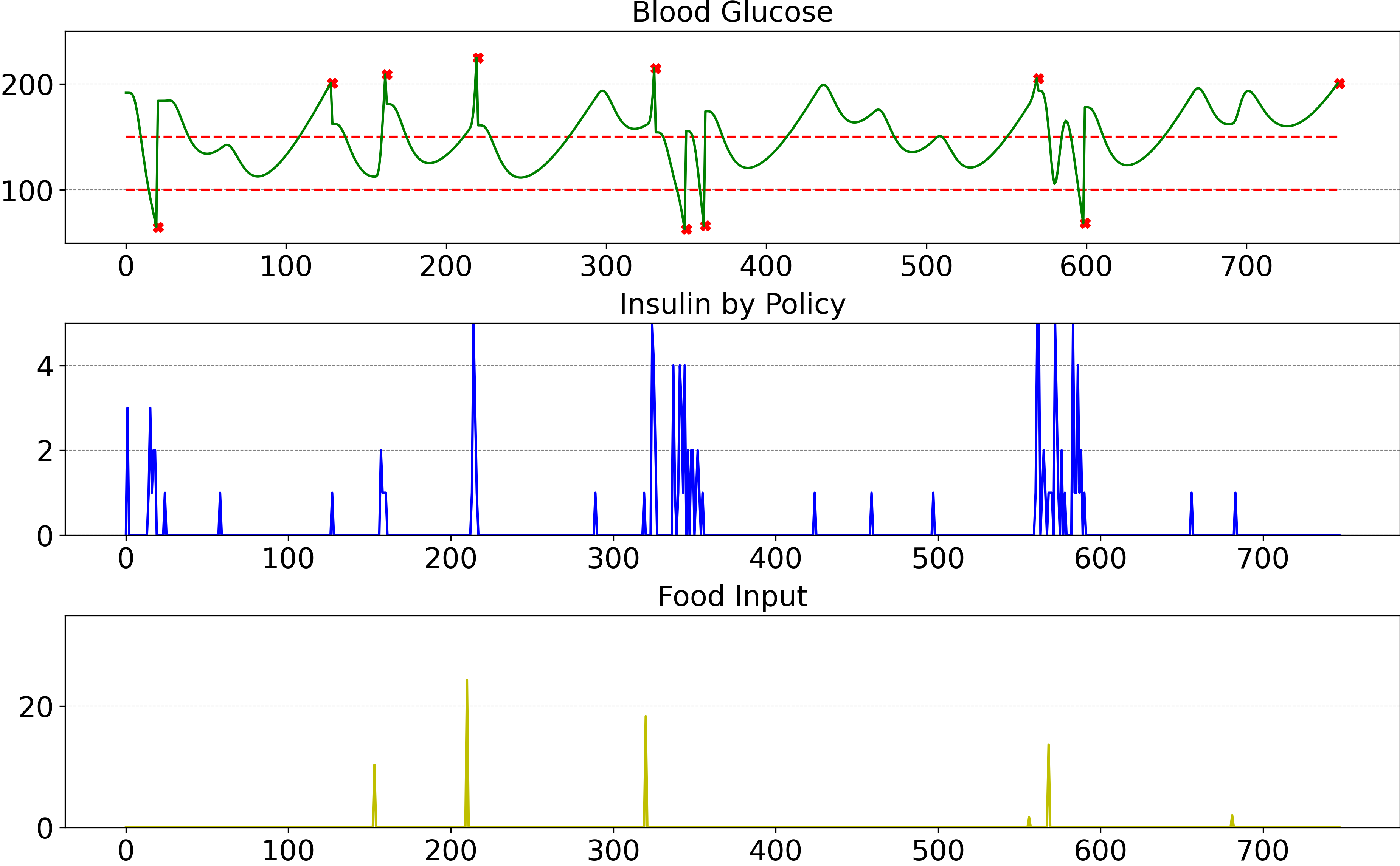}
  \caption{ADRQN}
%   \label{fig:eval_adrqn}
\end{subfigure}
\begin{subfigure}{.37\textwidth}
  \centering
  \includegraphics[width=0.95\linewidth]{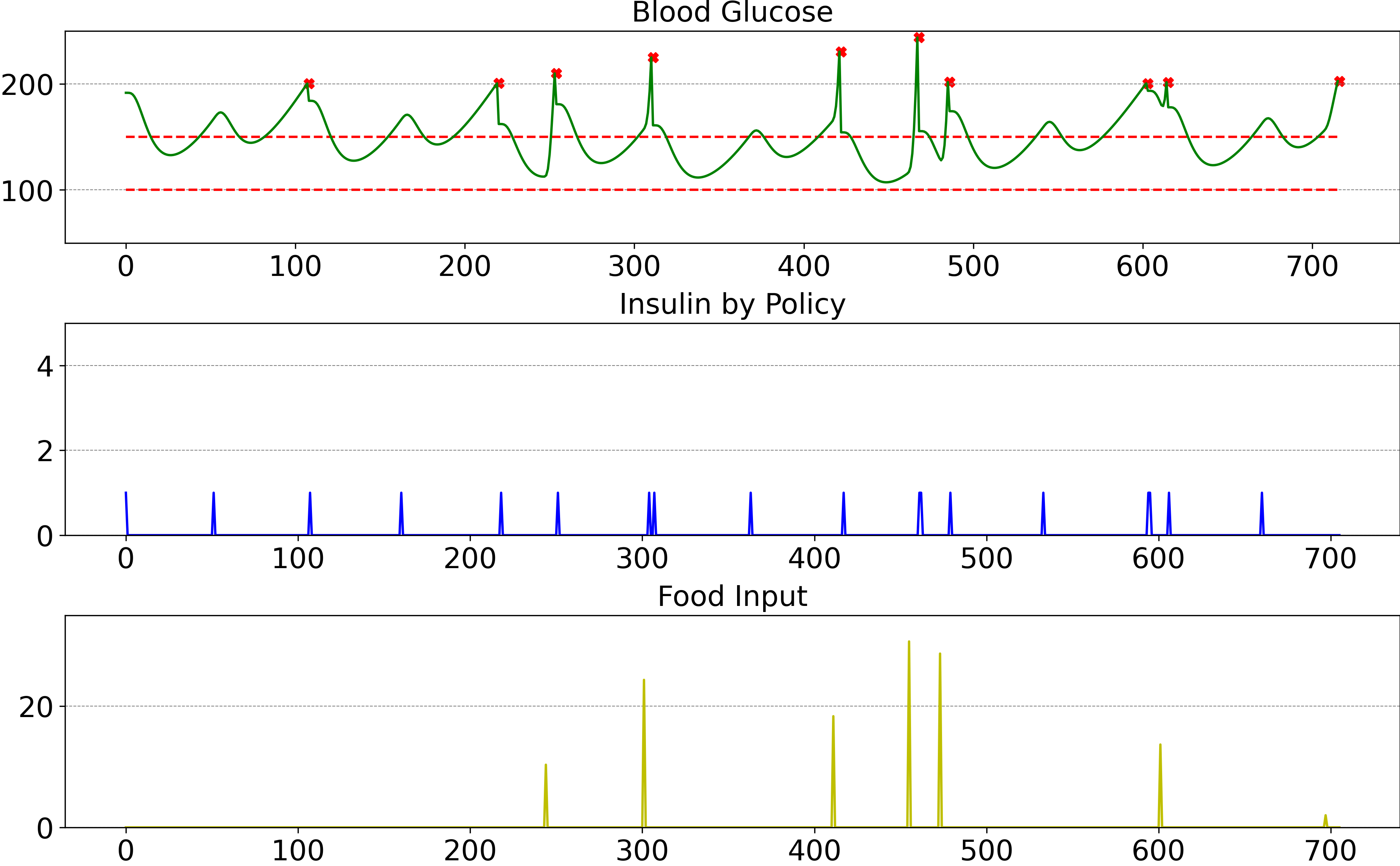}
  \caption{Effective-DQN}
%   \label{fig:eval_effdqn}
\end{subfigure}
\caption{\textbf{Policy visualization.} Trained greedy agents evaluated on 10 episodes with the same starting blood glucose. Red crosses mark the end of episodes. The insulin administered is as prescribed by the respective policy. In the blood glucose plot, the space between the two red dotted lines is the target zone. Effective-DQN appears to be conservative and thus does not allow the patient to get into hypoglycemia. The ADRQN agent gets into both hypo- and hyperglycemia.}
\label{fig:evaluate}
\end{figure*}

We present a qualitative assessment of the policies under comparison to better understand their behavior. Figure~\ref{fig:evaluate} displays the effect of insulin intake dictated by each policy, and food intake, on the blood glucose level over 10 evaluation episodes. Note that the food intake schedule is fixed across all agents and episodes, and that food occurrence might look different across different plots due to episodes terminating, in some cases, before mealtime. The starting blood glucose level is also the same in all cases, but from then on, the trajectories are generated based on each policy. The fixed policy (Figure~\ref{fig:evaluate}(a)) administers the same dose given the same blood glucose level, and hence it is not surprising to see all the episodes terminate in hypoglycemia due to the prolonged effect of the doses injected at higher blood glucose levels. The DQN agent (Figure~\ref{fig:evaluate}(b)) learns to administer a constant low dose of insulin due to the various penalties in the reward function. A probable interpretation is that: (1) The agent learns that not doing anything -- \ie setting the dose to $0$ --  makes the agent spend more time in hyperglycemia, and incurs penalties (2) Due to the inherent prolongedness of actions, higher doses quickly drive the agent into hypoglycemia, which also leads to penalties. Since the agent has no prior information about the prolonged action effects, it may interpret this penalty as indicating that choosing high doses is harmful. (3) There is also an additional penalty that comes from overdosing. Hence, the DQN agent chooses a middle ground of constantly administering the minimum non-zero dose available (dose = 1) to maximize its duration into the target zone. The ADRQN agent (Figure~\ref{fig:evaluate}(c)) is able to leverage insulin history through its state description and hence mostly administers new doses sparingly. As shown in the figure, it chooses all possible doses between 0 and 5. We observe frequent doses when blood glucose spikes around food intake, but in an attempt to quickly regulate the spike in blood glucose, the agent often ends up over-dosing. This leads to quick episode termination due to hypoglycemia right after food consumption. This is an undesirable behavior, evoking safety concerns, as in the clinical context, hypoglycemia is more fatal than hyperglycemia \citep{mccrimmon2010}, and requires immediate medical attention.
Finally, the Effective-DQN agent (Figure~\ref{fig:evaluate}(d)) appears to be conservative; like the DQN agent, it only injects the smallest possible dose when it decides to administer insulin. However, unlike the DQN agent, the Effective-DQN agent has information about the residual insulin in the blood from previous doses and hence learns when not to administer further doses. This is an important improvement over DQN, which closes the performance gap with ADRQN. It is interesting to note that the Effective-DQN agent also chooses to administer insulin around food intake when blood glucose spikes, but the dose magnitude remains small. A noteworthy behavior is that the Effective-DQN agent never ended an episode due to hypoglycemia. A potential explanation is that either administering a slightly higher does of 2 units or more frequently administering 1-unit doses might push the agent into hypoglycemia and eventually death.  This hypothesis could be verified in the future by training an agent with access to smaller insulin doses. Note however that this behavior is an observation, not a constraint, and hence not a guarantee on the policy. To have such guarantees one must add desired conservative properties as constraints, which we leave for future work.\looseness-1

% \paragraph{Sensitivity Analysis}
% Both history length in ADRQN and $\lambda$ in Effective DQN capture the prolonged-ness and hence are crucial hyper-parameters for restoring the Markov assumption. In this section we will analyze the respective model's sensitivity to these parameters. As shown in figure \ref{fig:ADRQN_history}, performance improves drastically from a history length of 5 to 10 after which it does not improve further. This saturation can be due to the constant model capacity across different horizons. We chose a history length of 15 over 10 as we found it to be more stable.

% \begin{figure}
% \centering
%   \centering
%      \includegraphics[width=\linewidth]{figs/effect_of_history_len_diabetes_lr0.001.png}
% \caption{\textbf{Effect of history length on ADRQN performance.} ADRQN agent does not improve beyond a history length 10, but we chose a more stable agent with history length 15. Solid line is mean and shaded region is standard deviation over 5 seeds. All results are smoothed with a window of size 100.}
% \label{fig:ADRQN_history}
% \end{figure}

% \begin{figure}
% \centering
%   \centering
%      \includegraphics[width=\linewidth]{figs/lamda_sensitivity.png}
% \caption{\textbf{Effect of $\lambda$ on Effective DQN performance.} Small values of $\lambda$ does not capture the prolongedness, and leads to poor performance. Values close to 1 leads to performance gain. Solid line is mean and shaded region is standard deviation over 5 seeds. All results are smoothed with a window of size 100.}
% \label{fig:lambda_sensitivity}
% \end{figure}

\section{Discussion}

% These solutions are tailored to specific sub-classes of POMDPs. The first step towards finding such solutions is to recognize different POMDP sub-classes, which can also trigger interest in associated illustrative environments and thus help push the frontiers beyond recurrent baselines. The results in the glucose control experiment reinforce this point. Our simple solution to handle PAE-POMDP not only outperforms a recurrent policy but also learns a safer policy, which is more suitable for a risk-averse application area such as healthcare.

\paragraph{Conclusion.} In this paper, we identified delay and prolongedness as a common roadblock for using RL in drug dosing. We defined PAE-POMDP, a sub-class of POMDPs in which action effects persist for multiple time steps, and introduced a simple and effective framework to convert PAE-POMDPs into MDPs which can be subsequently solved with traditional RL algorithms. We evaluated our proposed approach on a toy task and on a glucose control task, for which we proposed a clinically-inspired reward function for the glucose control simulator, which we hope will facilitate further research. Our quantitative results have shown that our proposed approach to convert PAE-POMDPs into MDPs is competitive and offers important advantages over the baselines. In particular, we have shown that by introducing domain knowledge on the prolongedness of action effects, we could build a solution capable of matching state-of-the-art recurrent policies such as ADRQN while being remarkably more time and memory efficient. Our qualitative results have further emphasized the benefits of the introduced Effective-DQN policy, which appeared to be more conservative than the recurrent policy, despite using the same reward function. \looseness-1

\paragraph{Limitations.} Our glucose control experiments are limited to a single virtual patient environment. Since humans exhibit similar profiles for prolonged action effects, similar performance on patients from other demographics is expected but should be evaluated. Although based on pharmacological literature we acknowledge that delay and prolongedness are present in any drug dosing scenario, we did not find other simulated environments to test our hypothesis. Hence, our experiments at the moment are limited to a single drug effect simulation. Although this work is only in the context of drug dosing, prolonged action effects seem to be present in other application areas such as robotics, where the decay assumption may not be justified. The recurrent solution is still the most general solution in such cases. This work does not explore those applications. \looseness-1

\paragraph{Future work.} In this work, we have only considered discrete action spaces. Exploring continuous actions might lead to more interesting insights  in the future. Delay and prolongedness appear across all drug dosing scenarios, so we hope future RL applications of autonomous drug dosing will consciously account for them. More broadly, future work includes coming up with more generalized solutions for action superposition that accommodate qualitative actions and are suitable for application areas beyond drug dosing.\looseness-1

% The resulting policy appears to circumvent fatal hypoglycemic conditions altogether by avoiding over-dosing and hence is likely to be safer than a recurrent policy for the glucose control task. 

\section*{Acknowledgements}
The first author is supported by IVADO PhD fellowship. We are grateful to Mila IDT (specially Olexa Bilaniuk) for their extraordinary support, and Emmanuel Bengio for the productive discussions.
% The first author would also like to acknowledge Mont Royal for being their late night lungs through the difficult Covid years.

\bibliography{aaai23}

% \section{Reference}
\label{sec:reference_examples}
% \nobibliography*

% Use \bibliography{yourbibfile} instead or the References section will not appear in your paper

% \bigskip
% \noindent Thank you for reading these instructions carefully. We look forward to receiving your electronic files!

\appendix

\section{Sources of Partial Observability in the Glucose control task}

\begin{figure}[H]
  \centering
  \includegraphics[width=\columnwidth]{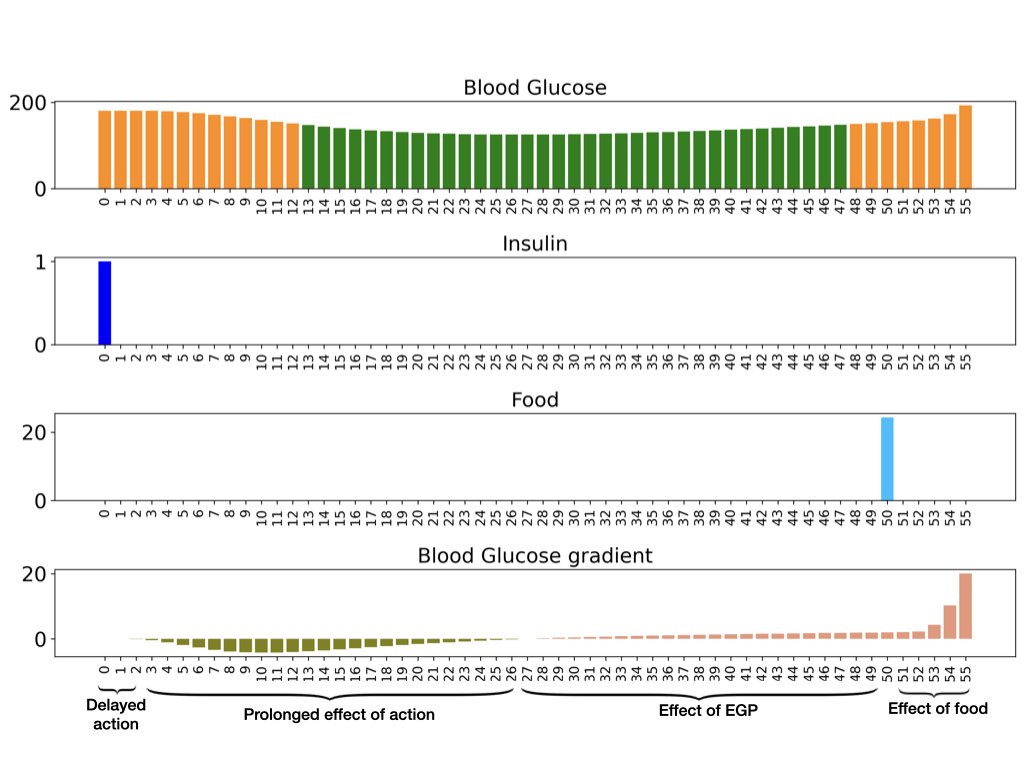}
  \caption{\textbf{Explaining glucose control environment.} Insulin has a delayed and prolonged effect on the blood glucose levels, which can also increase due to Endogenous Glucose Production (EGP) under fasting. x-axis is time steps, y-axis is respective units. In the Blood Glucose subplot, the color green represents the target zone and orange the hyperglycemic zone. In the Blood Glucose Gradient subplot, the color olive-green represents a reduction and salmon-pink an increase in the blood glucose level.}
  \label{fig:diabetes_pomdp_source}
\end{figure}

As shown in figure \ref{fig:diabetes_pomdp_source}, there are multiple sources of partial observability in this environment:\\
\textbf{1) Prolonged Effect:} The effect of insulin on blood glucose is prolonged over multiple time steps. A high dose of insulin can quickly terminate the episode due to hypoglycemia, thereby reducing the episode length. The duration of the effect depends on an individual's insulin sensitivity and is proportional to the magnitude of the dose for lower doses of insulin.\\
\textbf{2) Constant Delay:} The insulin administered at the sub-cutaneous level takes some time to reach the bloodstream and be effective. This delay is constant for an individual.\\
\textbf{3) Food:} An individual can consume food at any point and hence the blood glucose can increase unexpectedly. The information of whether a meal has been consumed is not part of the state space and is an element of surprise, thus another source of partial observability.\\
\textbf{4) Endogenous Glucose Production:} When fasting, the human body produces glucose by breaking down glycogen stored in the liver to meet the brain's energy demand \citep{sharabi2015}.\\
\textbf{5) Sensor Noise:} The blood glucose measurements by the sensors of Continuous Glucose Monitor (CGM) can have a little noise, making the CGM readings slightly different from the actual blood glucose levels. However, to keep the experiments simple, for the experiments in this paper we made noise as an identity function, making the CGM readings same as the blood glucose.

\section{Simglucose simulator changes}
We made a few changes to the simglucose simulator:

\begin{enumerate}

     \item \textbf{Starting state:} The original simulator starts all the episodes with the same blood glucose value of 150, which is not favorable for exploration. Therefore, we randomize the starting blood glucose in the range of 150 to 195. Specifically, we modified the \texttt{reset()} method of \texttt{T1DPatient()} class in \texttt{simglucose/patient/t1dpatient.py}.
     
     \item \textbf{Episode termination:} The original simulator terminates the episodes if the blood glucose is below 70 or higher than 350. We redefine the termination criteria to 70 and 200. We made this change to keep the search space small for the experiments presented in this paper.
    
    \item \textbf{Removing Noise:} We removed noise by changing \texttt{CGM = BG} in the \texttt{measure()} method of the \texttt{CGMSensor()} class found in \texttt{`simglucose/sensor/cgm.py'} file.
    
    \item \textbf{Reward function:} We implement a reward function as described in the main paper. The reward function is shared with the code.
    
    \item \textbf{Action space:} The original simulator allows insulin dose of upto 30 units. Therefore, the action space is continuous between 0 and 30. We reduce the action space to discrete values of 0 to 5 to simplify the task for the paper. 
    % We added gym wrapper to discretize the action space and reduce it to 6 actions (0-5)
    
    \item \textbf{Environment and Episode:} we randomly select the patient adult009 as the environment to train the RL agents presented in this paper and assume that we do not have access to the transition dynamics. We slightly modify the environment to redefine episodes. Our episodes begin uniformly in the blood glucose range of (150, 195) and terminate if the blood glucose is greater than 200 (hyperglycemic death) or less than 70 (hypoglycemic death). We define \textit{hypoglycemia} as the condition when blood glucose ranges from 70 to 100, \textit{target zone} as 100 to 150 and \textit{hyperglycemia} as 150 to 200. We also reduce the action space to discrete doses of insulin of up to 5 units. The objective is to effectively administer insulin to maximize the time spent in the target blood glucose zone.
\end{enumerate}

For more details or the exact implementation, please refer to the copy of the updated simulator code added to the supplementary package. We will make them public, on acceptance of the paper. 

\section{Experimental Details}

\subsection{Architecture}
Both move block and glucose control task use the same architecture, but only differ in the number of neurons.
% Both the task use the same network architecture, but differ at the number of neurons at each layer.
\begin{itemize}
    \item \textbf{State embedding:} For both the simulator (simglucose and MoveBlock), state is a 1 dimensional continuous value. We use a fully connected layer of 16 neurons with linear activation to learn a state-embedding. We train this state-embedding together with the rest of the Q-network. Although weights are not shared, we use the same state-embedding architecture for all three of our models.
    
    \item \textbf{Action embedding:} Action is a 1 dimensional discrete value. We use a fully connected layer with linear activation to learn an action-embedding. Effective DQN and ADRQN use the same architecture for the action embedding. They don't share weights and we train them together with the rest of the Q-network and state-embedding if applicable. MoveBlock uses 32 neurons and the glucose control task uses 16 neurons to encode the actions.
    
    \item \textbf{DQN:} DQN passes the state-embedding through a fully connected layer and ReLU activation before passing it through a fully connected output layer.
    
    \item \textbf{Effective DQN:} Effective DQN concatenates the state and action embeddings before passing them through a fully connected layer with Relu activation followed by a fully connected output layer.
    
    \item \textbf{ADRQN:} The original ADRQN architecture uses convolutional layers, but since our application has low dimensional input, we replace the convolutional layers with linear layers similar to \cite{peschl2020}. It concatenates the state and action embeddings before passing them through a LSTM unit followed by a fully connected output layer.
    
MoveBlock uses 512 neurons and glucose control uses 256 neurons for the hidden layer.
    
\end{itemize}

\subsection{Hyper-parameters}

We selected the hyper parameters by grid-search and present the final set of selected hyper-parameters for the function approximation experiments in Table \ref{table:hyperparam-table}.

We use a learning rate of 0.001 for tabular q-learning and 0.005 for effective q-learning on the tabular toy task. We choose a lambda value of 0.99 for the effective q-learning. For both the algorithms, we use an epsilon greedy policy that decays exponentially from 0.9 to 0.05 over a span of 10K episodes.

\begin{table}
  \caption{Hyperparameters used to train agents on the glucose control task}
  \label{table:hyperparam-table}
  \centering
  \resizebox{\columnwidth}{!}{
  \begin{tabular}{lccc}
    \toprule
    % \multicolumn{2}{c}{Part}                   \\
    % \cmidrule(r){1-2}
     \textbf{Hyper-parameter}  &  \textbf{DQN} &  \textbf{ADRQN} &  \textbf{Eff-DQN}\\
    \midrule
    \multicolumn{1}{c}{Glucose Control Task}                   \\
    \cmidrule(r){1-4}
    Learning Rate & 1e-6 & 0.001 & 0.001 \\
    State embedding dimension & 16 & 16 & 16 \\
    Action embedding dimension & N/A & 16 & 16 \\
    history length & N/A & N/A & 15 \\
    lambda & N/A & N/A & 0.99 \\
    hidden layer dimension & 256 & 256 & 256 \\
    seed & 1-5 & 1-5 & 1-5 \\
    replay buffer size & 100K & 100K & 100K \\
    batch size & 512 & 512 & 512 \\
    Explore & 1000 & 1000 & 1000 \\
    Target network update rate & 1 & 1 & 1 \\
    % \midrule
    % \multicolumn{1}{c}{MoveBlock}                   \\
    % \cmidrule(r){1-4}
    % Learning Rate &  0.0001 & 0.001 & 1e-5 \\
    % State embedding dimension & 16 & 16 & 16 \\
    % Action embedding dimension & N/A & 32  & 32  \\
    % history length & N/A & N/A & 1 \\
    % lambda & N/A & N/A & 0.3 \\
    % hidden layer dimension & 512 & 512 & 512 \\
    % seed & 1-5 & 1-5 & 1-5 \\
    % replay buffer size & 100K & 100K & 100K \\
    % batch size & 256 & 256 & 256  \\
    % Explore & 1000 & 1000 & 1000 \\
    % Target network update rate & 1 & 1 & 1 \\
    \bottomrule
  \end{tabular}}
\end{table}

\subsection{Implementation details}
We used the DQN implementation from the PyTorch DQN tutorial \cite{paszke2021}
% \footnote{https://pytorch.org/tutorials/intermediate/reinforcement\_q\_learning.html}
of PyTorch forum. We implemented effective DQN on top of this. The ADRQN implementation published with the main paper was in C++ and Caffe. Therefore, we have used an open-source PyTorch implementation \cite{peschl2020}.
% \footnotemark[\ref{adrqn}]. 
We self implemented the driver and helper codes and shared them with this supplementary package.

\subsection{Qualitative Analysis Data generation}
\begin{table}
  \caption{Data generation for the Qualitative Analysis}
  \label{table:eval-table}
  \centering
  \resizebox{\columnwidth}{!}{
  \begin{tabular}{ccccc}
    \toprule
    seed & Starting Blood Glucose & \multicolumn{3}{c}{Ending Blood Glucose}\\
    % \multicolumn{2}{c}{Part}                   \\
    \cmidrule(r){3-5}
    & & \textbf{DQN} & \textbf{ADRQN} & \textbf{Effective-DQN}\\
    \midrule
    100 & 191.57  & 65.60 & 65.01 & 200.26\\
    110 & 184.02  &  63.19 & 200.66 & 200.49\\
    120 & 162.14 & 69.08 & 208.86 & 209.82\\
    130 & 180.80 & 62.17 & 224.56 & 224.90\\
    140 & 160.92 & 68.61 & 214.56 & 230.39\\
    150 & 154.27 & 66.07 & 63.10 & 243.63\\
    160 & 155.43 & 67.66 & 66.16 & 201.68\\
    170 & 174.21 &  60.08 & 205.03 & 200.42\\
    180 & 193.46 & 66.84 & 68.64 & 201.52\\
    190 & 177.90 &  61.25 & 200.37 & 202.49\\
    \bottomrule
  \end{tabular}}
\end{table}

\begin{figure*}[ht]
    \centering
    \includegraphics[width=\textwidth]{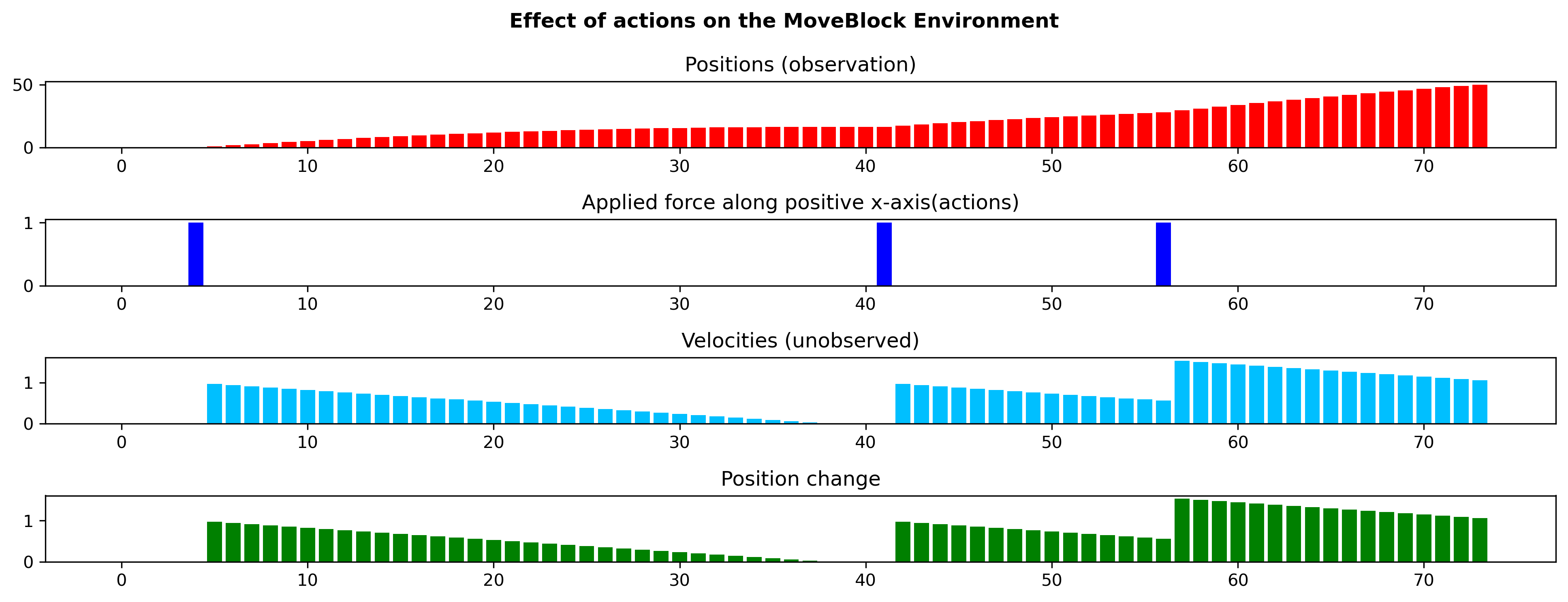}
    \caption{\textbf{Prolonged and Additive Action Effect in MoveBlock Environment.} \emph{Prolonged Effect:} Position (observed variable) does not change initially in the absence of force (action). Unit force is exerted at t=3 followed by no force till t=40. But since velocity (unobserved variable) keeps on changing under the previously exerted force until friction brings it to zero, position keeps on changing too in the absence of any further force. \emph{Additive Effect:} Forces exerted at t=41 and t=57 exhibit additive nature.}
    \label{fig:pae-env}
\end{figure*}

For the quantitative assesment, we trained 5 agents corresponding to seeds 1 to 5 for DQN, ADRQN and Effective-DQN. Out of the trained 5 agents we randomly chose the agents with seed 3 for all the policies. For the qualitative assessment, we fix 10 starting blood glucose levels and use them to generate trajectories for each policy under evaluation. In table \ref{table:eval-table}, we present the seeds used to generate the episodes. For sanity check, we also present the starting blood glucose levels corresponding to these 10 seeds. The trajectories depend on the policy in use. We report the end of trajectory blood glucose levels at the end of each episode.

\section{Prolonged action effect on \textit{MoveBlock} toy environment}
Figure \ref{fig:pae-env} demonstrates the effect of prolonged actions on the states on the toy environment. We observe that actions have a prolonged effect on the latent variable velocity, which in turn affects the observed variable position. In the toy environment we could control the prolonged effect by changing the co-efficient of kinetic friction. We used a coefficient of 0.003 for the environment used in the experiments. The reward function is same as the the Mountain Car environment:
\begin{equation}
\label{eq:moveblock_reward}
r(s, a) = 
\begin{cases}
    100 &\text{if $s \in S_{goal}$}\\
    - a^2 * 0.1 &\text{otherwise}\\
\end{cases}
\end{equation}

\end{document}